\documentclass[letterpaper, 10 pt, conference]{ieeeconf}  

\IEEEoverridecommandlockouts                              

\usepackage{graphics} 
\usepackage{epsfig} 
\usepackage{mathptmx} 
\usepackage{times} 
\usepackage{amsmath} 
\usepackage{amssymb}  
\usepackage{url}
\usepackage{mars}
\title{\LARGE \bf
Towards Generation and Evaluation of \\Comprehensive Mapping Robot Datasets 
}

\extrafloats{100}

\author{Hongyu Chen$^{1}$, Xiting Zhao, Jianwen Luo, Zhijie Yang, Zehao Zhao, Haochuan Wan,\\ Xiaoya Ye, Guangyuan Weng, Zhenpeng He, Tian Dong, and S\"oren  Schwertfeger$^{1}$
	\thanks{$^{1}$All authors are with ShanghaiTech University
		{\tt\small <chenhy3, soerensch>@shanghaitech.edu.cn}}%
}

\begin{document}

%
%


\marsPublishedIn{Accepted for:} 		

\marsVenue{IEEE Conference on Robotics and Automation (ICRA) 2019 workshop}

\marsYear{2019}

\marsPlainAutors{ Hongyu Chen$^{1}$, Xiting Zhao, Jianwen Luo, Zhijie Yang, Zehao Zhao, Haochuan Wan,\\ Xiaoya Ye, Guangyuan Weng, Zhenpeng He, Tian Dong, and S\"oren  Schwertfeger$^{1}$}


\marsMakeCitation{Towards Generation and Evaluation of Comprehensive Mapping Robot Datasets}{IEEE Press}


\marsIEEE{}


\makeMARStitle

\maketitle
\thispagestyle{empty}
\pagestyle{empty}

\begin{abstract}
This paper presents a fully hardware synchronized mapping robot with support for a hardware synchronized external tracking system, for super-precise timing and localization. We also employ a professional, static 3D scanner for ground truth map collection. Three datasets are generated to evaluate the performance of mapping algorithms within a room and between rooms. Based on these datasets we generate maps and trajectory data, which is then fed into evaluation algorithms. The mapping and evaluation procedures are made in a very easily reproducible manner for maximum comparability. In the end we can draw a couple of conclusions about the tested SLAM algorithms.
\end{abstract}

\section{INTRODUCTION}
Localization and mapping are essential robotic tasks and are often solved together in a Simultaneous Localization and Mapping (SLAM) system \cite{cadena2016past}. SLAM systems have to be evaluated regarding their performance. But this is not a trivial task. 

Often, ground truth robot paths are used to measure the quality of the SLAM system, since it is assumed that a good localization will result in good maps. In \cite{WulfGroundTruthEval2007} and \cite{Kuemmerle2009onMeasuring} the ground truth paths are compared with the paths estimated by the SLAM algorithms. A metric for measuring the error of the manually corrected trajectory of datasets is available to the public in \cite{Burgard-SLAMcompGraphOfRelations-IROS09}. Recently, Zhang and Scaramuzza have provided a tutorial and software for quantitative trajectory evaluation \cite{zhang2018tutorial}. 

If ground truth paths are not available, the maps can be used to evaluate the quality of the mapping system. For that image similarity methods \cite{MapQuality-RoboCup08} and pixel-level feature detectors \cite{PellenzMappingAndMapScoring2008, Lakaemper2009VirtualScans} can be applied to the maps, but have their limitations, because maps often have errors like structures appearing more than once due to localization errors. More high level features like barrels are used for evaluation in \cite{Schwertfeger2010Fiducials,SSRR11-Schwertfeger-Fiducials} and also in 3D maps in \cite{Schwertfeger2015Fiducials3D}.

Another approach is to capture the topology of the maps and use the matches for comparison \cite{Schwertfeger2015_Topo_AuRo}. There are also evaluation methods that don't rely on ground truth data. In \cite{Newmann-MarkovFields-MapQuality08}  suspicious and plausible arrangements of planes in 3D scans are detected and the map is evaluated accordingly.

\begin{figure}[t]
	\centering 
	\includegraphics[width=0.8\linewidth]{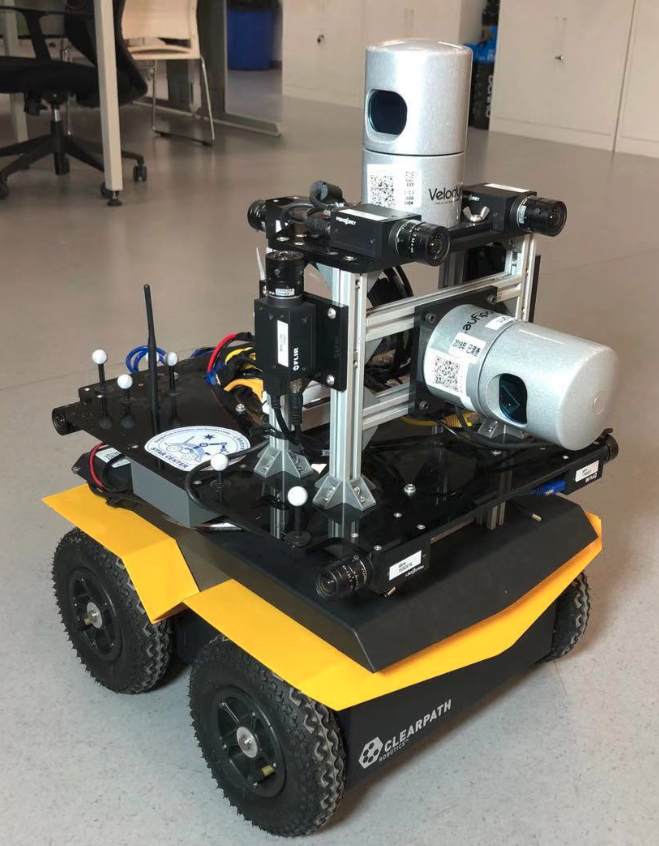}  
		\caption{The MARS Mapper robot with sensors, that is used in this paper. }
	\label{fig:robot_platform}   
\end{figure}

\begin{figure*}[bt]
	\centering
	\includegraphics[width=1.0\linewidth]{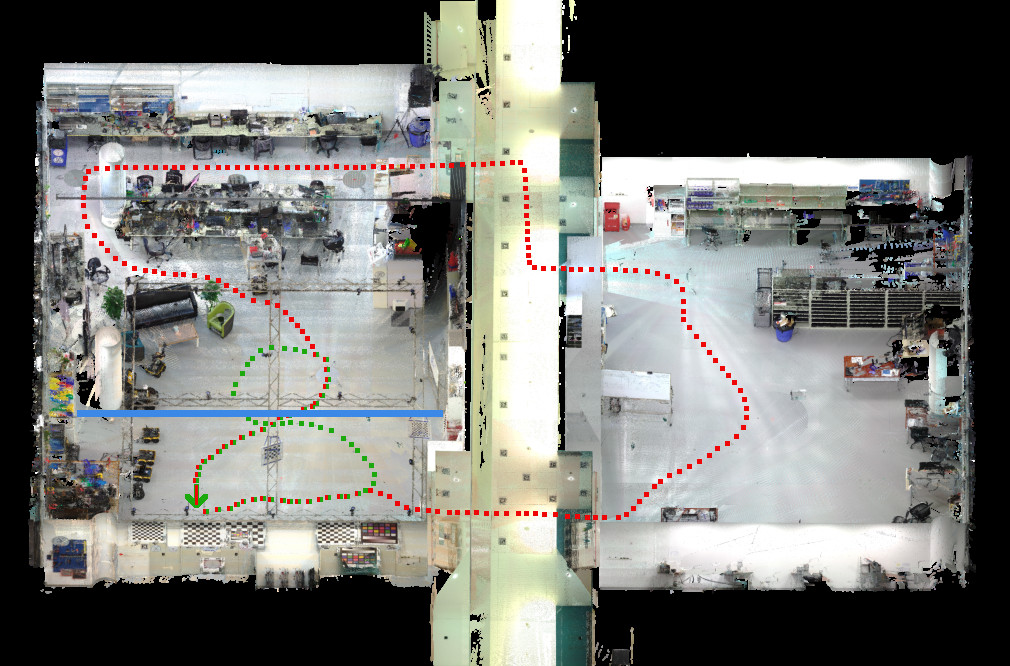}  
	\caption{Top-view of the ground-truth faro point cloud (480mill points) with the MARS-8 path (green), the MARS-Loop and MARS-NoLoop paths (red) 
as well as the location of the curtain (blue) added.}
	\label{fig:paths}
\end{figure*}

Often, simulations are used for the evaluation of SLAM systems \cite{scrapper-NavigationMetrics-SSRR07}. A great factor for errors in SLAM systems is the noise of the sensors. It is thus essential to accurately simulate the noise and errors in the simulation, a task which is not simple. Datasets of 3D sensor measurements and ground truth poses for benchmarking the performance of SLAM algorithms are provided in \cite{Sturm2012-BenchmarkRGBDslam} and \cite{Handa2015-3DSlamDataset}. The ground truth information has been obtained using a tracking system and by creating the data in a simulation, respectively.

\par
In this paper we propose to use an advanced mapping system with sensors that are hardware-synchronized to an external tracking system to collect data for benchmarking datasets. We believe this approach is a valuable supplement to SLAM evaluation using simulations, because it ensures real sensor noise and real locomotion, vibrations and other factors that are difficult to accurately simulate. Using the tracking system we gather ground truth localization information. But we believe that it is also important to evaluate the mapping performance, especially for visual SLAM. So we also collect ground truth map information using a professional, static 3D scanner (Faro Focus 3D). 

We provide three datasets. One purely within the tracking system, especially also for evaluating mapping performance. Additionally we have two longer, very similar datasets, that start and end in the tracking system. One can do loop closing at the end, the other cannot (no overlapping sensor data). Using these datasets we aim to evaluate the loop-closing and scan-matching performance.

We provide ROS launch files to automatically generate the maps from the datasets and to evaluate them by comparison to ground truth. Our results are thus fully reproducible.

The key contributions of this paper are thus:
\begin{itemize}
  \item Presentation of a hardware-synchronized advanced mapping robot with external tracking.
  \item Generation of three datasets for mapping and SLAM evaluation.
  \item Providing reproducible evaluations of standard mapping software based on the datasets.
\end{itemize}

This work is work-in-progress and we will provide more mapping solutions and evaluation systems based on the datasets with the final version of this paper, if it gets accepted.

The reminder of the paper is organized as follows: Section \ref{sec:mapper} describes the mapping robot and the other hardware and procedures for dataset generation. In Section \ref{sec:datasets} we describe the three datasets we provide. The reproducible evaluation of those datasets explained in Section \ref{sec:eval}, followed by the conclusions in Section \ref{sec:conclusions}.

\section{Mapping System}
\label{sec:mapper}

Our mapping system consists of a Clearpath Jackal robot with an upgraded power supply and computer (Intel Core i7-6770k CPU, Raid 0: 3x Samsung 850 EVO 500G). The mainboard supports eight independent USB 3.1 ports, which are mainly used for the cameras. The robot is collecting data from the following sensors:

\begin{itemize}
  \item Nine 5MP wide-angle color cameras (FLIR Grasshopper3 GS3-U3-51S5C-C) with wide-angle lenses (82$^{\circ}$ x 61$^{\circ}$), 10Hz (4 stereo pairs: front, left, right, up; one back-looking camera)
  \item Two Velodyne HDL-32E 3D Laser scanners, 10Hz (one horizontal, one vertical; both in dual-return mode)
  \item IMU Xsense MTi-300, 200Hz
  \item Robot odometry
  \item Optitrack tracking system (21 Prime 13 cameras, 30Hz)
\end{itemize}

We compress the camera images with JPEG quality 90. Due to CPU speed limitations we can not store much more than 10Hz for the 9 cameras, so we chose to collect the images with the same frequency as the Velodynes. This results in a total storage bandwidth of about 170 MB/s.

\subsection{Synchronization}
Especially with respect to quality evaluation, but also for mapping systems in general, it is important to synchronize all the sensor data. We use an Asus Tinker Board with a quad core 1.8 GHz ARM Cortex-A17 processor to provide hardware synchronization. The Tinker Board serves as our reference time. It is triggering the cameras and the tracking system with 10 Hz and the Xsens IMU with 200Hz. But hardware triggering of sensors is only half of the job: Afterwards the data from the cameras arrives on PC in the Jackal robot at different times, due to USB and CPU scheduling issues. We thus need to be able to associate the hardware triggers with the actual data and make use of this info in the software. Thus, for every trigger it generates, the Tinker board also sends a ROS timestamp to the PC, which is collected in the bagfile. In a post-processing step we then match the sensor data with this time-stamp and then correct the time-stamps of the data. Afterwards all data (e.g. all images and the tracking data and the IMU information) that was triggered together will have exactly the same time-stamp. 

The tracking system is also triggered by the Tinker board, but it is then increasing the frequency from 10Hz to 30Hz. For that the robot is physically connected to the tracking system via an ethernet cable when inside the systems camera view. Before leaving or entering the tracking system we manually (un)plugged this cable, briefly stopping the data-collection for this. To avoid having a second ethernet cable from the tracking system to the robot, we collect the tracking system's (which is running on Windows) data on a separate Ubuntu PC and merge the bagfiles later. Before each run, the time of the PC, the Jackal PC and the Tinker Board are updated via NTP from the router of our lab.

Since the Velodyne is a rotating sensor, it cannot be triggered. Instead it time-stamps its messages using GPS pps and NMEA data. The Tinker Board is providing fake GPS data with its own time to the Velodynes, such that their data arrives at the Jackal PC already with the correct time stamp. 

\subsection{Calibration}
\begin{figure}[tb]
		\centering
		\includegraphics[width=1.0\linewidth]{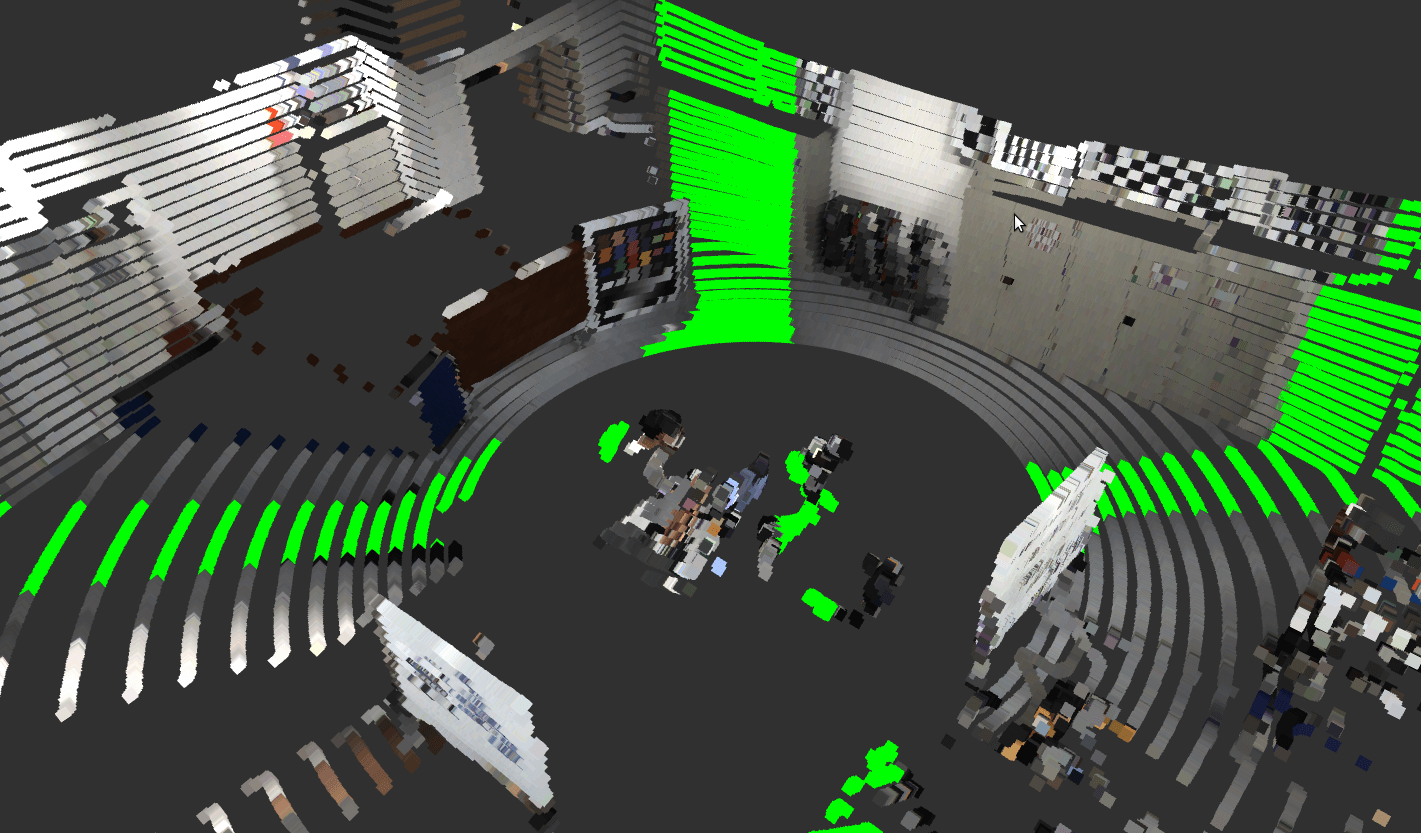}  
	\caption{ A 3D Lidar scan colored by all the $7$ horizontal cameras. The transformations between each camera and 3D lidar are acquired from global optimization result. All the green points represent areas where no camera is overlapping with the point cloud. }
   \label{fig:coloredpoints}
\end{figure}

Our MARS Mapper robot (MARS is the acronym of our Mobile Autonomous Robotic Systems Lab) is fully calibrated. Intrinsic calibrations are acquired using the known methods. The extrinsic calibration of the sensors (i.e. their poses) are gathered by pair-wise calibration of various, also heterogeneous sensor pairs (4x stereo cameras, 32x non-overlapping cameras, 13x lidar to camera, 1x lidar to lidar, 9x tracking system to camera) and then minimizing the error using G2O \cite{kummerle2011g}. Figure \ref{fig:coloredpoints} shows a Velodyne scan where the points that are within the field of view of one of the 7 horizontal cameras are colored. 

Because we calibrated the sensors with the tracking system, we can ready use the tracking system results as the pose of the robot. The tracking system reports an error of the collected poses of less than 1.5mm.

\section{Datasets}
\label{sec:datasets}

The datasets were collected in the Mobile Autonomous Robotic Systems Lab (MARS Lab) of ShanghaiTech University, and have thus MARS in their name:
\begin{itemize}
  \item \textbf{MARS-8}: A short (23m) figure eight driven by the mapping robot with continuous tracking information. For basic SLAM evaluation and evaluation of mapping performance.
  \item \textbf{MARS-Loop}: A medium length (77m) mapping run, starting in the tracking system in the MARS lab, leaving the lab and re-entering it through a different door, finally entering the tracking system again and finishing at the start pose. For evaluation of basic loop closing performance.
  \item \textbf{MARS-NoLoop}: The MARS lab is devided into two parts by two curtains (10cm apart; along the center of the tracking system). The robot follows the same path as MARS-Loop, except that it stops a little earlier (because the curtains are in the way). The robot starts and ends in the same tracking system. No loop closing is possible between the start and end of the dataset, because there is almost no overlap between the areas. 
\end{itemize}

Figure \ref{fig:paths} shows the paths of the robot in the different datasets: green for MARS-8 and red for MARS-Loop and MARS-NoLoop. The approximate paths we followed are also marked on the ground and are thus visible in some of the collected camera images (black for MARS-8 (and MARS-Loop where they overlap) and white for MARS-Loop). MARS-NoLoop is following MARS-Loop, except stopping earlier. Almost nothing in the environment was changed between the robot data collections nor for the Faro scans.

\begin{figure}[tb]
	\centering
	\includegraphics[width=0.9\linewidth]{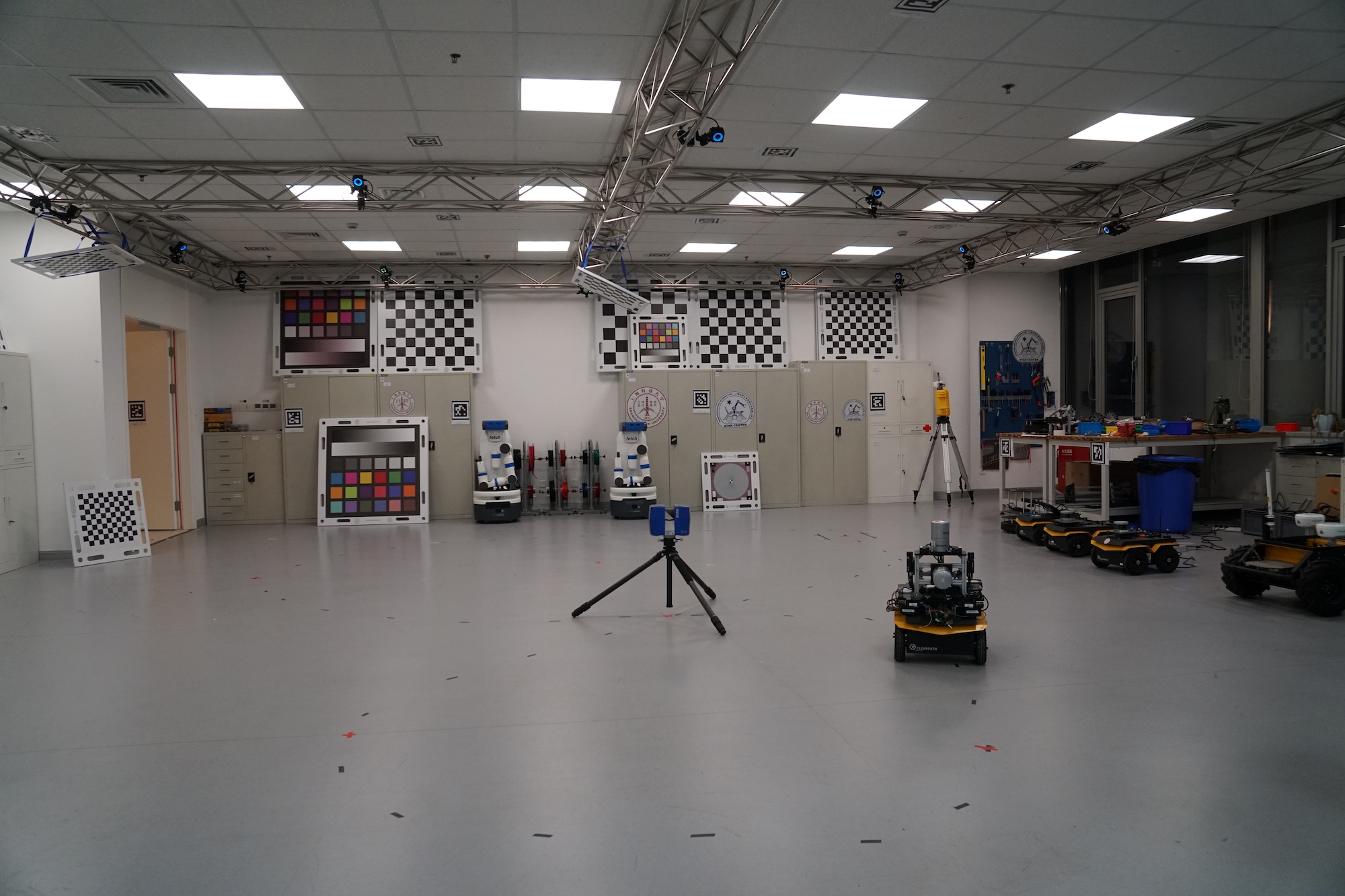}  \\ \vspace{2mm}
	\includegraphics[width=0.9\linewidth]{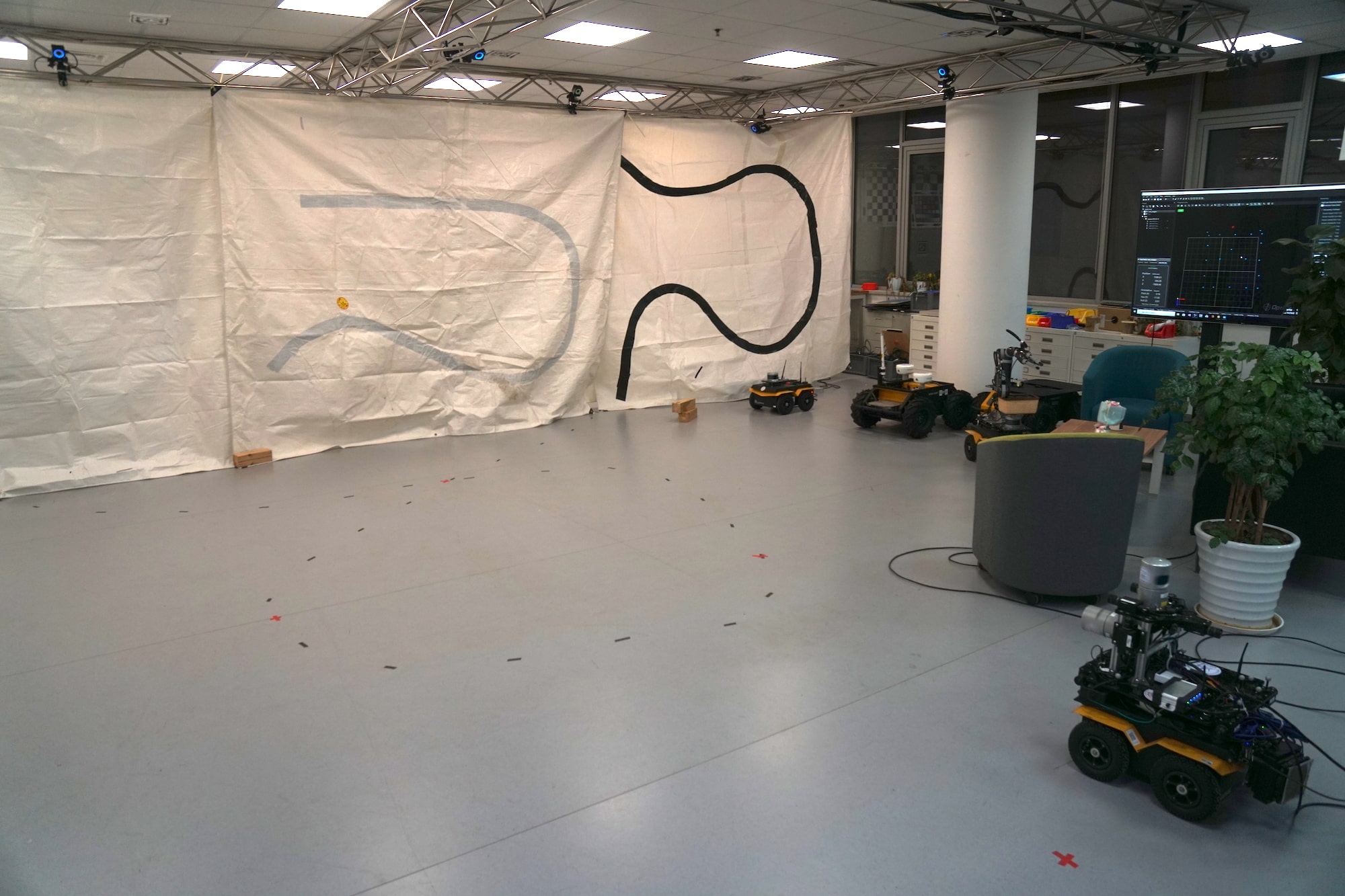}  
	\caption{The robot and the FARO scanner in the MARS lab on top. Below, the same area with the curtain (made with tarp) for map MARS-NoLoop.   }
	\label{fig:lab}
\end{figure}

Figure \ref{fig:paths} is actually the complete point cloud (480 million points) from the 18 FARO scans we collected (each about 27 million points). We used the FARO Scene software to register the scans. It reported an average error between the scan points of 1.2mm. This is an excellent value and much smaller than the expected sensor noise. The FARO data can thus serve as ground truth for map comparison. The approximate positions of the Faro scans are marked with red crosses on the ground. Most of the scans were taken at a hight similar to the horizontal Velodyne (61 cm). Figure \ref{fig:lab} shows the MARS Lab with the markings on the floor, the Faro scanner and the robot. It also shows the curtain for MARS-NoLoop.

We also placed several checkerboards in the lab. Additionally we have many April tags distributed on the ceiling and, in the MARS Lab, also on the walls. In the future we plan to evaluate how well those can be used for localization evaluation of SLAM algorithms. For good measure we also placed other cool robots of the MARS Lab as well as a small living-room arrangement with sofa, plants and TV in the scene.

Dataset statistics:
\begin{itemize}
  \item \textbf{MARS-8}: 16.4GB; 99 seconds  
  \item \textbf{MARS-Loop}: 50.7GB; 290 seconds
    \item \textbf{MARS-NoLoop}: 54.8GB; 315 seconds
  \item \textbf{MARS-8-Sample}: 500MB, 3 seconds 
\end{itemize}

The datasets are available online \footnote{\url{https://robotics.shanghaitech.edu.cn/datasets/MARS-Dataset}}. We also provide a very short and small sample dataset from within MARS-8.

\section{Evaluation}
\label{sec:eval}

Scientific results should be reproducible. Since we provide the dataset, we also want to give the reader the possibility to re-create the exact same map (barring differences caused by randomized SLAM algorithms). We are thus providing ROS launch files (start scripts) that generate the maps and other needed information (e.g. the path estimated by the SLAM algorithm). Furthermore, we also want to make it as easy as possible for the user to then evaluate the result, so we provide the according tools on the dataset website.

We apply the following mapping methods to our dataset:
\begin{itemize}
  \item 2D Grid Mapping (converting the horizontal Velodyne scan in a 2D LRF message; 5cm resolution):
  \begin{itemize}
    \item Hector Mapping \cite{KohlbrecherMeyerStrykKlingaufFlexibleSlamSystem2011}
    \item Cartographer \cite{hess2016real} 
    \item GMapping \cite{grisetti2007improved}
  \end{itemize} 
  \item 3D Point Cloud Mapping (with horizontal Velodyne):
  \begin{itemize}
    \item BLAM \footnote{\url{https://github.com/erik-nelson/blam}}
    \item BLAM with both Velodynes
    \item Cartographer (for final version)
  \end{itemize}
  \item visual SLAM:
  \begin{itemize}
    \item ORB2 \cite{blair2001design}
    \item RTAB-Map mono \cite{labbe2018rtab} (for final version)
    \item RTAB-Map stereo (for final version)
  \end{itemize}
\end{itemize}

If needed we modify those algorithms to output the time-stamped path data as a text file. The trajectory estimated by the SLAM algorithms is then compared to the trajectory of the tracking system. Of course this only works for the parts of the datasets that were collected inside the tracking system - other parts are omitted (and "jumped over"). We use the software provided by \cite{zhang2018tutorial} for the evaluation.

\newcommand\blaSize{0.3}

\newcommand\hecSize{0.3}

\begin{figure*}

	\centering

	\includegraphics[width=\blaSize\linewidth]{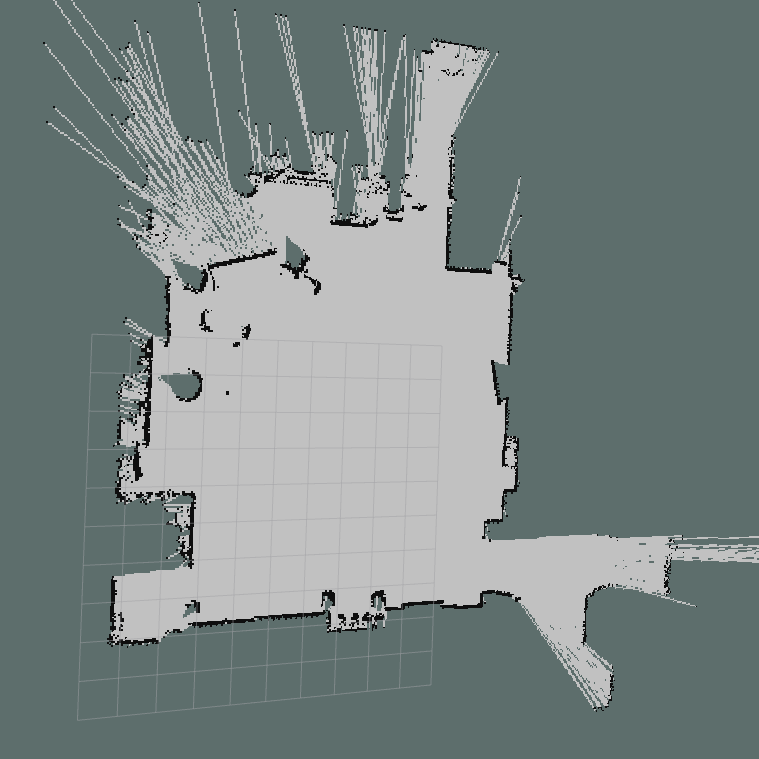}  
	\includegraphics[width=\blaSize\linewidth]{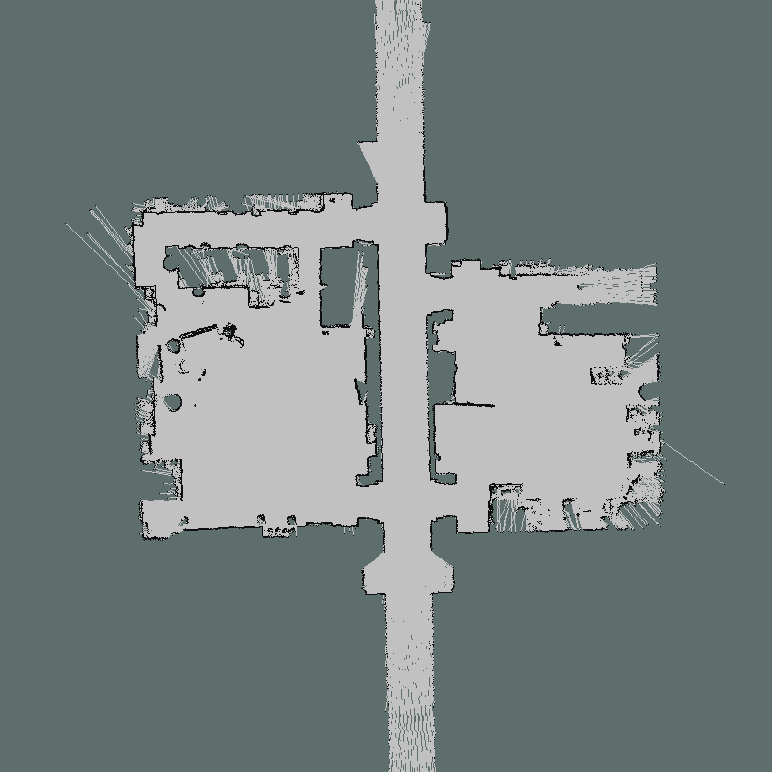}  
	\includegraphics[width=\blaSize\linewidth]{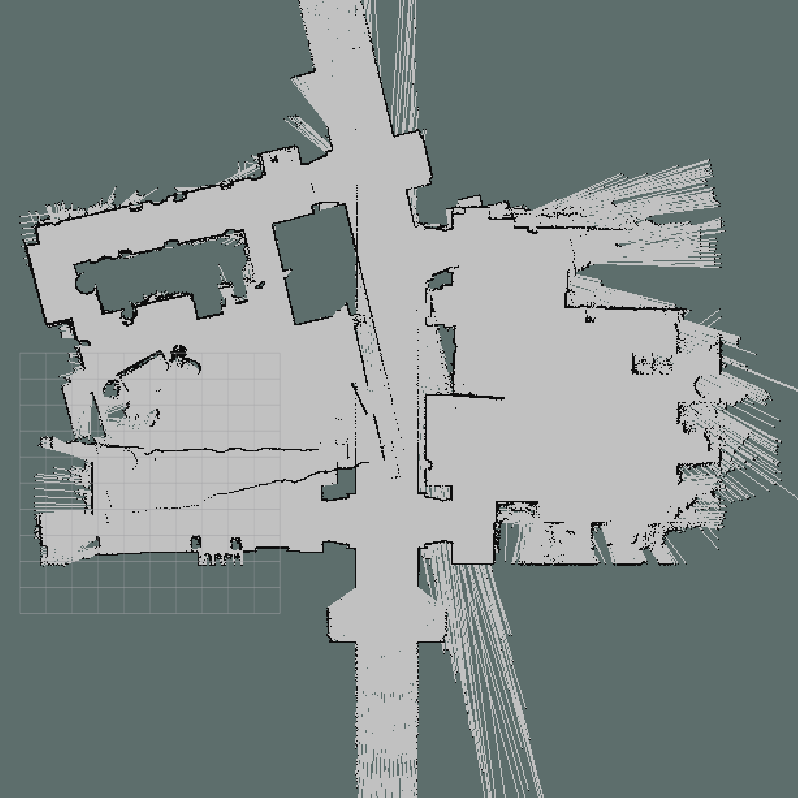}  \\ \vspace{1mm}

	\includegraphics[width=\blaSize\linewidth]{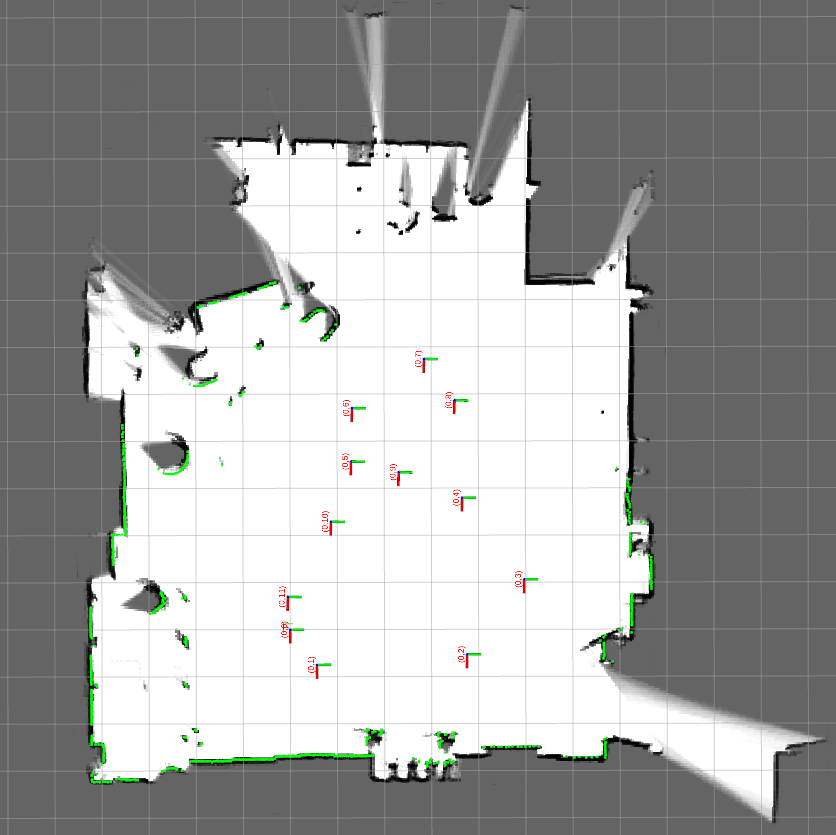}    
	\includegraphics[width=\blaSize\linewidth]{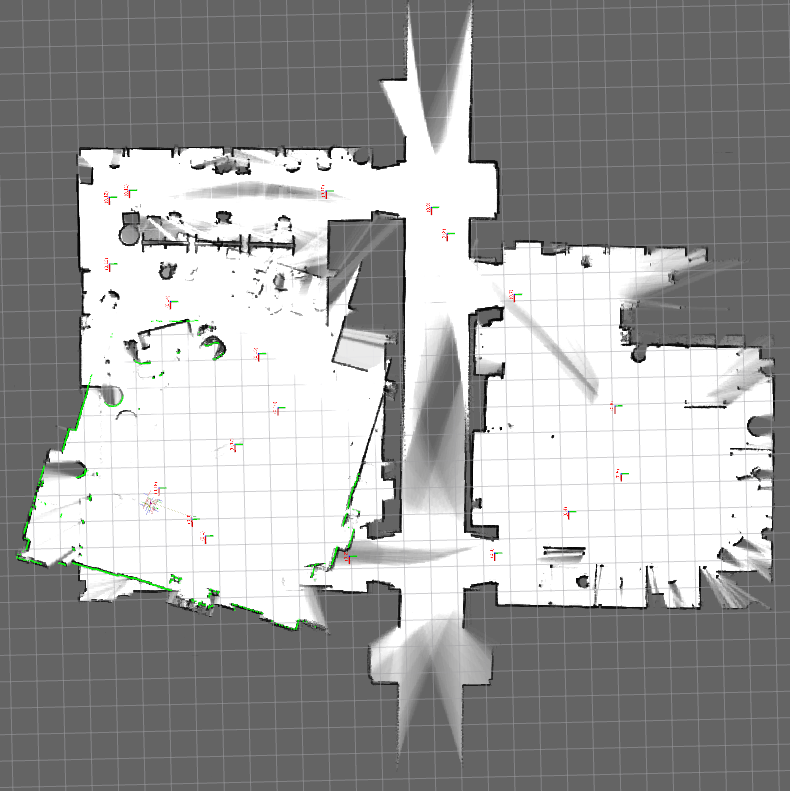}    
	\includegraphics[width=\blaSize\linewidth]{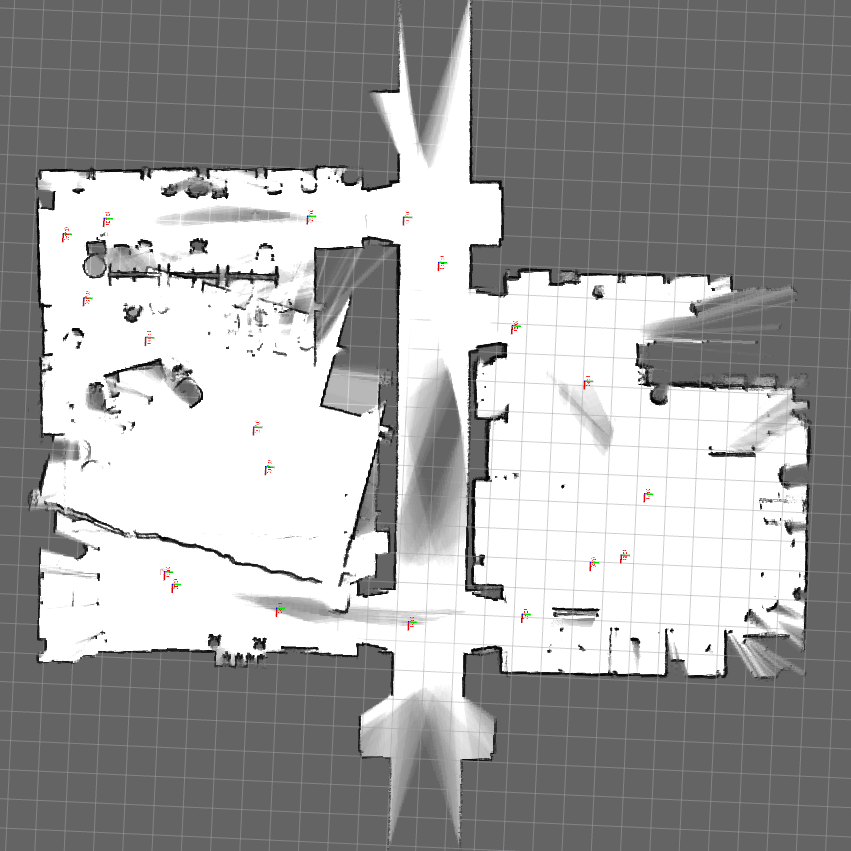}     

	\caption{The 2D grid maps created by Hector Mapping on top and cartographer below, on maps MARS-8, MARS-Loop and MARS-NoLoop, respectively. }
	\label{fig:maps_2d}


	\centering
	\includegraphics[width=\hecSize\linewidth]{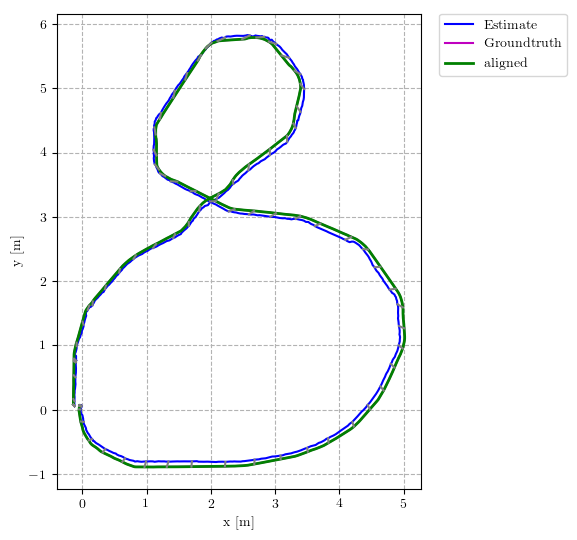}  
	\includegraphics[width=\hecSize\linewidth]{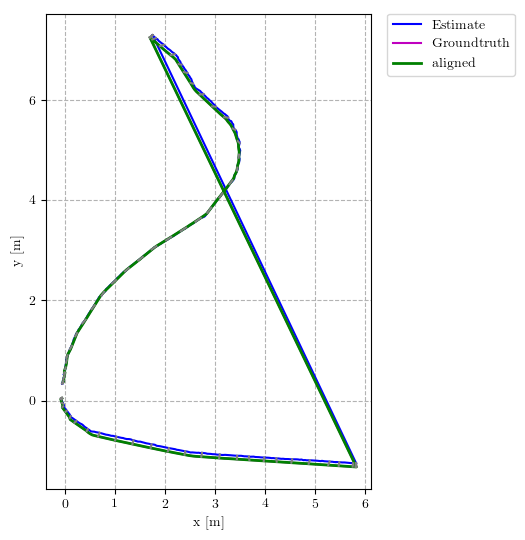}  
	\includegraphics[width=\hecSize\linewidth]{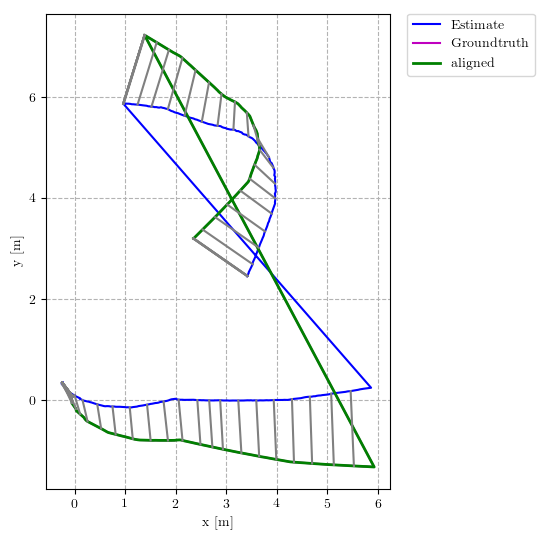}  \\ \vspace{1mm}

	\includegraphics[width=\hecSize\linewidth]{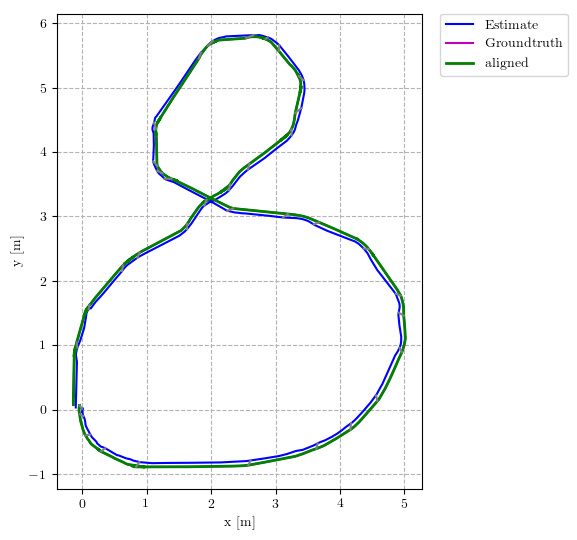}  
	\includegraphics[width=\hecSize\linewidth]{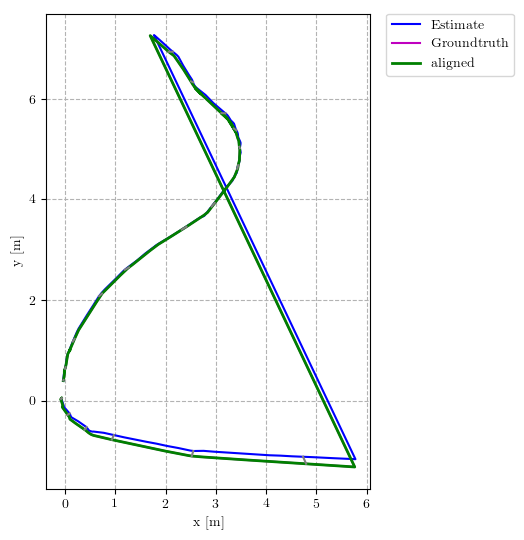}  
	\includegraphics[width=\hecSize\linewidth]{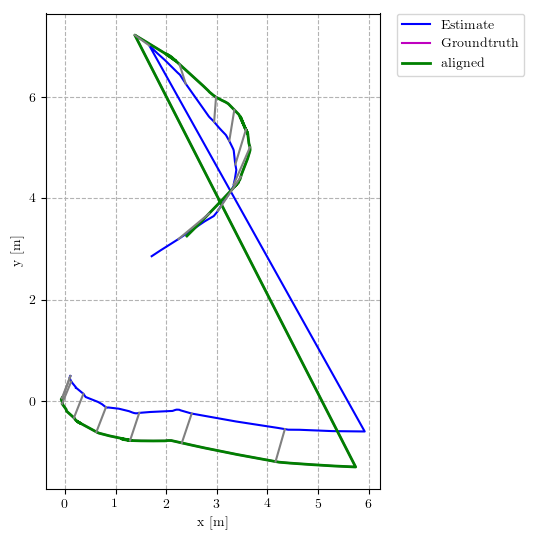}  
	\caption{The trajectories of the robot compared to the tracking system trajectories. On top Hector Mapping on MARS-8, MARS-Loop and MARS-NoLoop, respectively. Cartographer below.}
	\label{fig:error_2d}
\end{figure*}

Figure \ref{fig:maps_2d} shows the 2D grid maps from Hector Mapping and cartographer. We see that both algorithms have problems when coming back into the MARS lab and no loop closing is possible. Cartographer even has a broken map in MARS-Loop. Figure \ref{fig:error_2d} shows the error in the trajectories between Hector (top) and cartographer (bottom). Again, note the jump from the bottom right corner to the top left in MARS-Loop and MARS-NoLoop: This is where the robot left the tracking system and later re-entered it. We can see that the shown error correlates nicely with the perceived map quality of Figure \ref{fig:maps_2d}. Figure \ref{fig:comparison} quantifies the error of Figure \ref{fig:error_2d} in a diagram. It shows absolute errors of 10cm for Hector MARS-8 and -Loop, but values round 1m error for the broken MARS-NoLoop. 


\newcommand\bluSize{0.3}

\newcommand\blamSize{0.3}

\begin{figure*}[tb]
	
	\centering
	\includegraphics[width=\bluSize\linewidth]{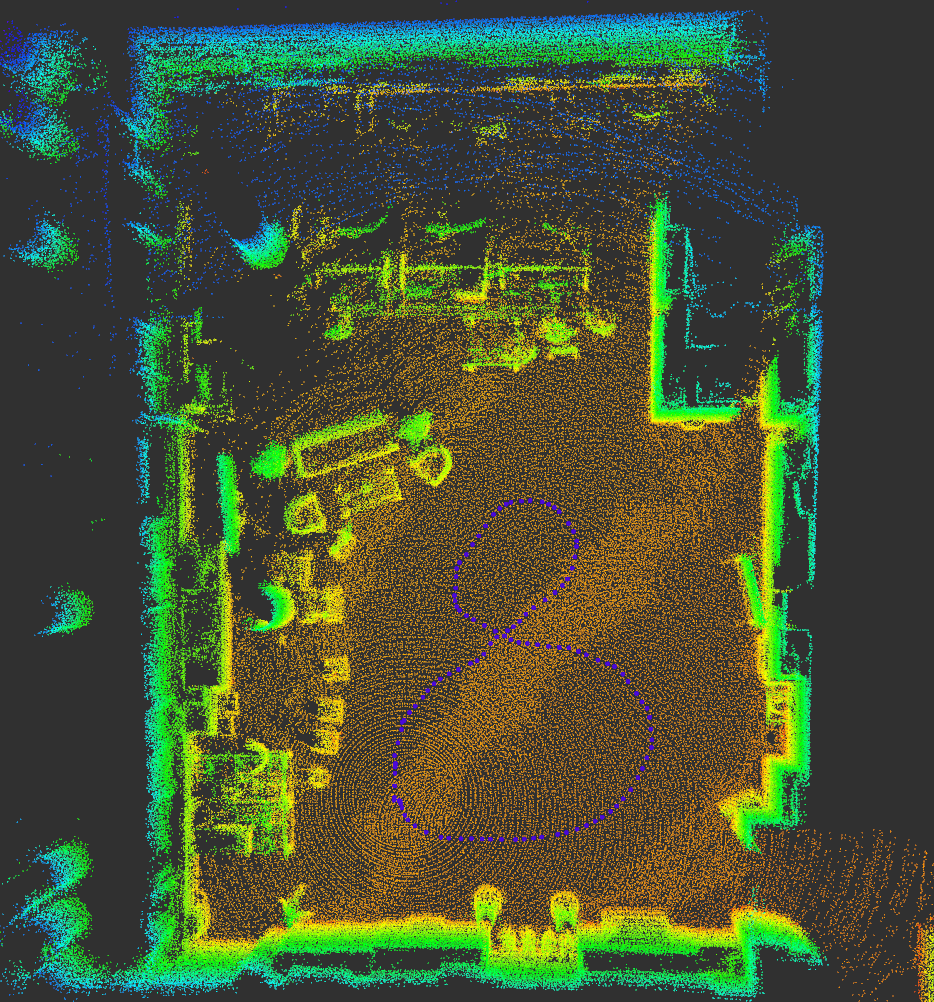}  
	\includegraphics[width=\bluSize\linewidth]{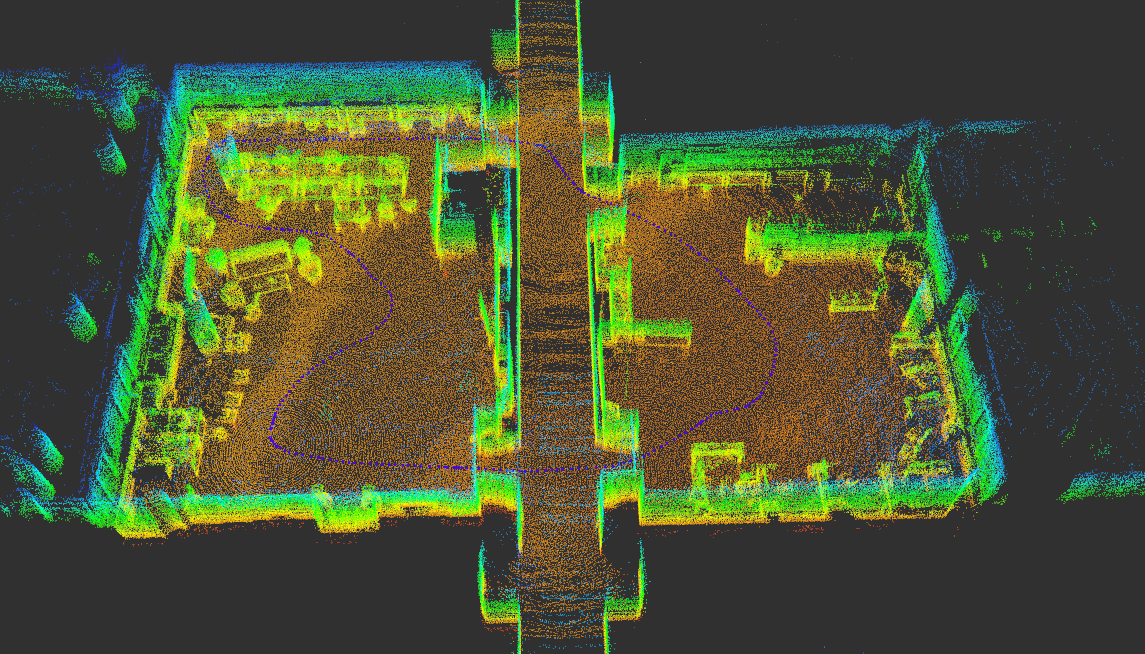}  
	\includegraphics[width=\bluSize\linewidth]{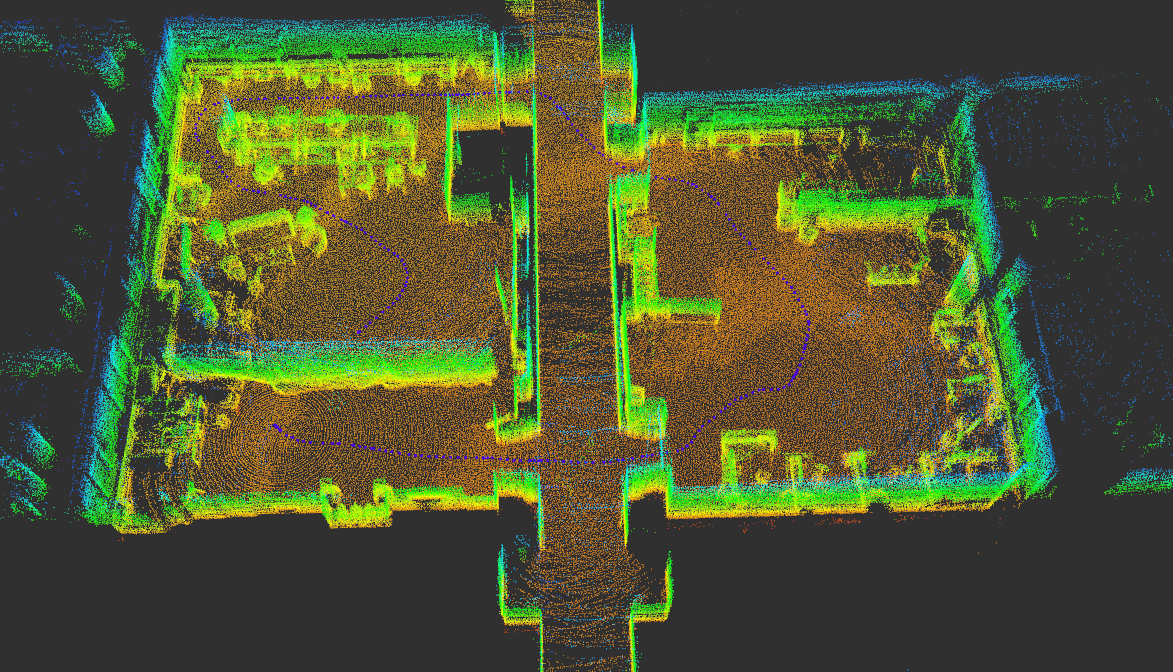}  

	\caption{3D maps generated by BLAM using the horizontal Velodyne for MARS-8 (left), MARS-Loop (middle) and MARS-NoLoop (right). }\vspace{6mm}
	\label{fig:blam}

	\includegraphics[width=\bluSize\linewidth]{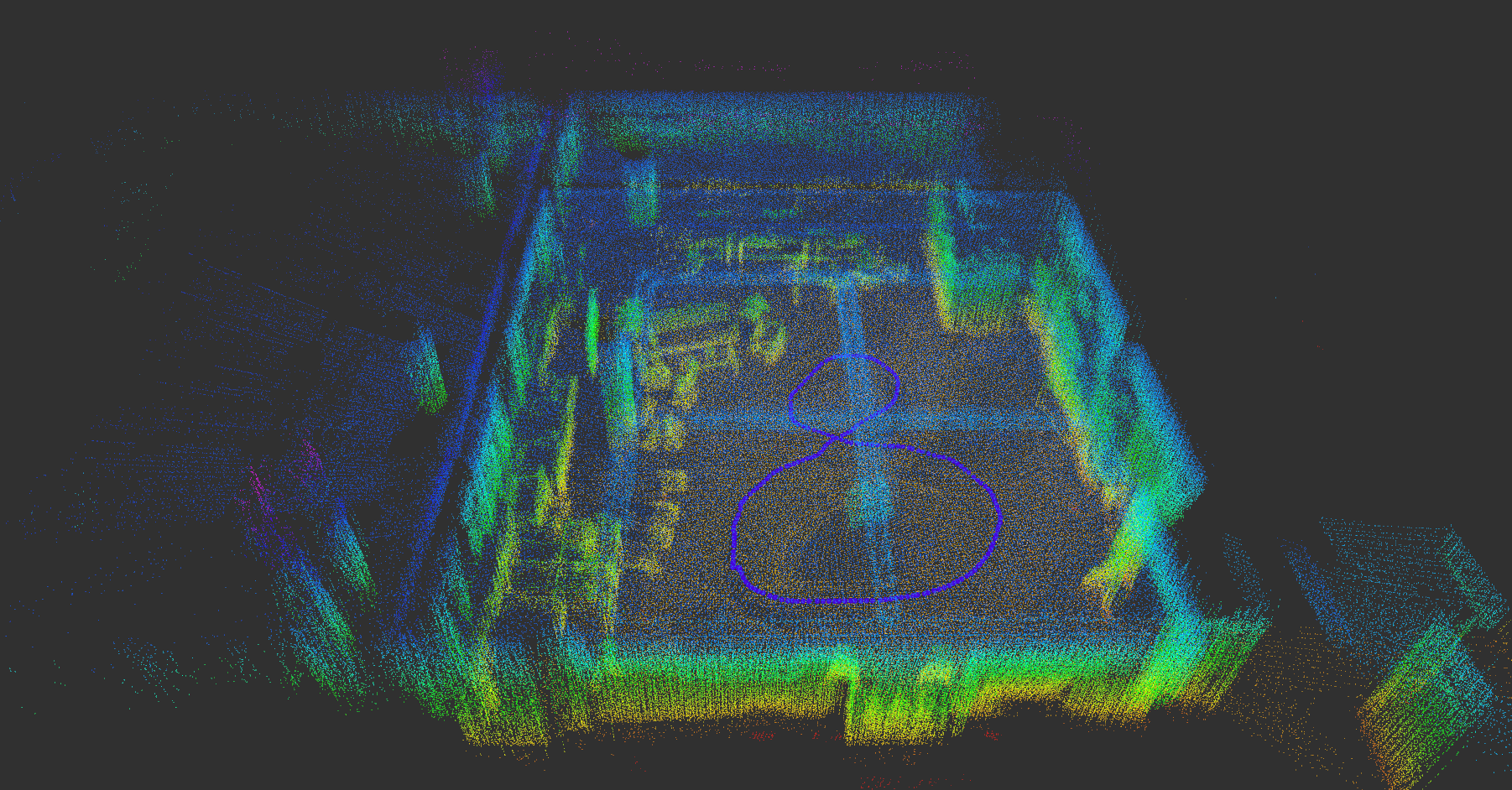}  
	\includegraphics[width=\bluSize\linewidth]{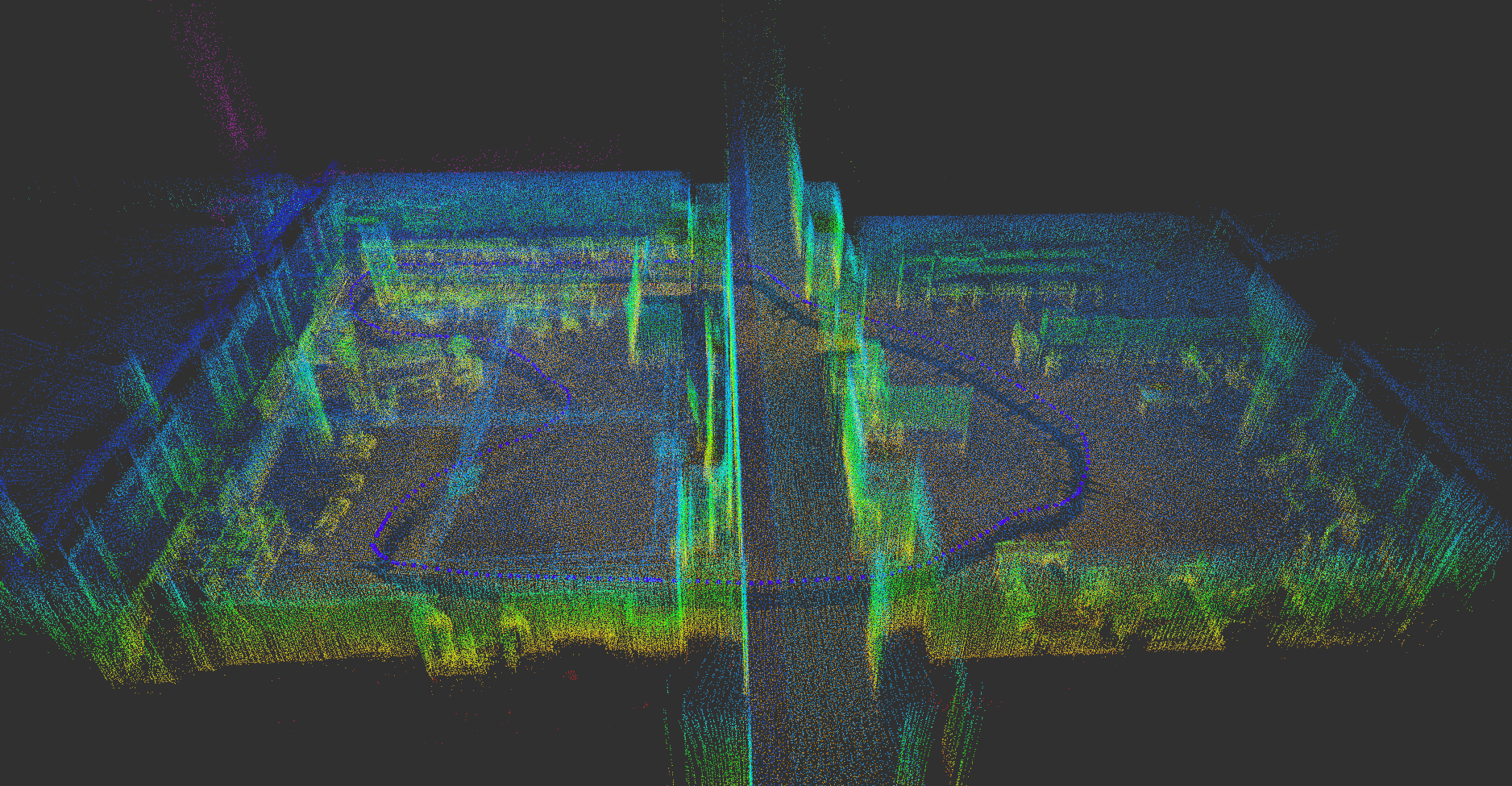}  
	\includegraphics[width=\bluSize\linewidth]{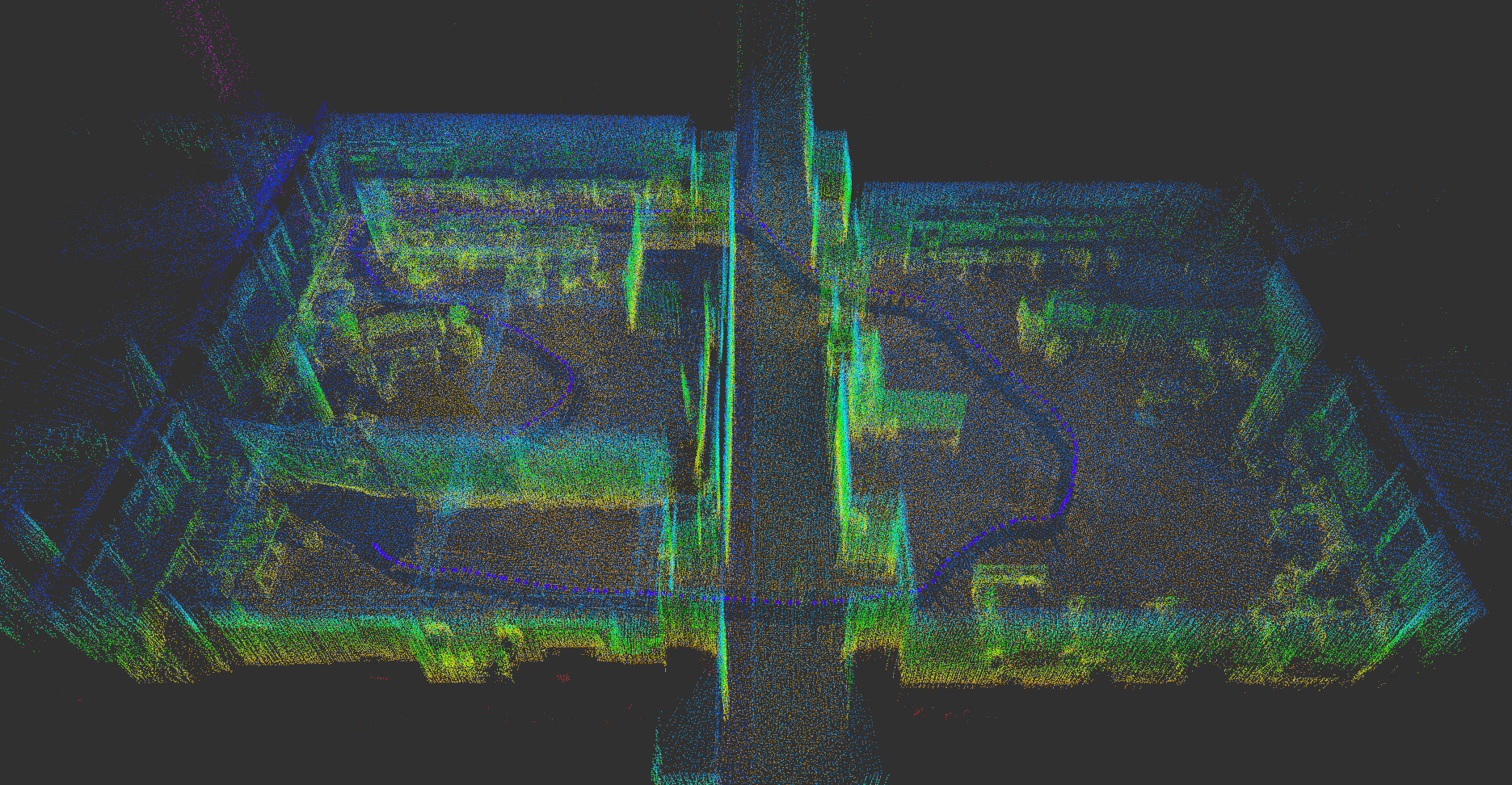}  
	
	\caption{3D maps generated by BLAM, rendering only the vertical Velodyne for MARS-8 (left), MARS-Loop (middle) and MARS-NoLoop (right).  }
	\label{fig:blam_v}
	
\end{figure*}

\begin{figure*}[tb]
	\centering

	\includegraphics[width=\blamSize\linewidth]{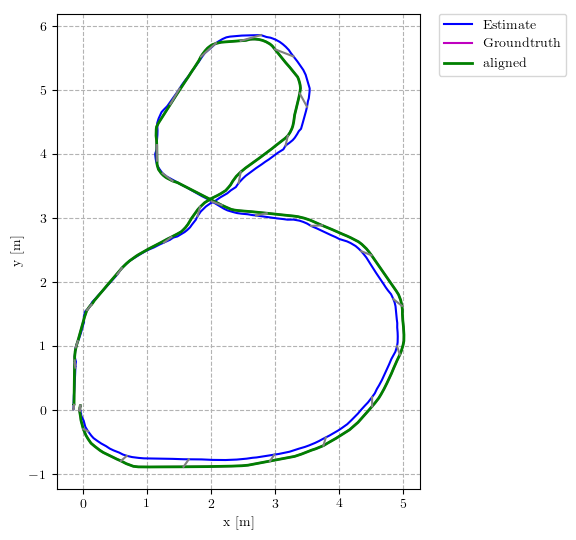}  
	\includegraphics[width=\blamSize\linewidth]{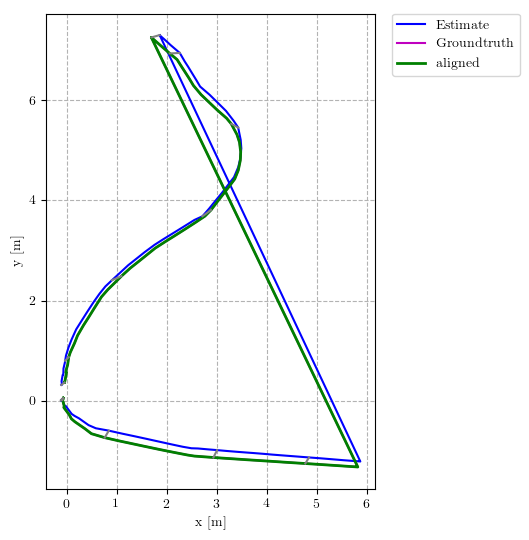}  
	\includegraphics[width=\blamSize\linewidth]{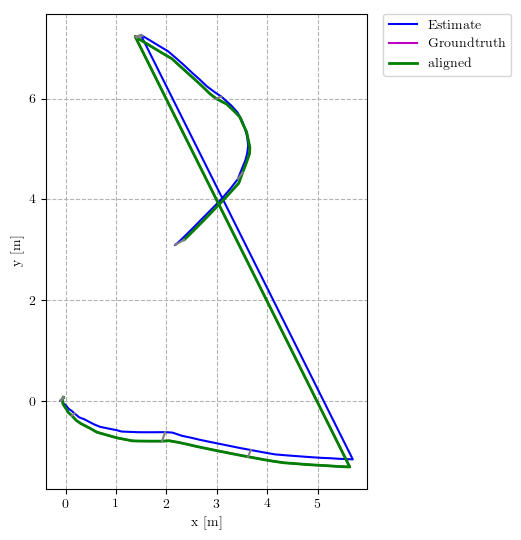}  
	\caption{Comparison of ground truth trajectory and BLAM for MARS-8, MARS-Loop and MARS-NoLoop, respectively.}
	\label{fig:blam_error}

\end{figure*}

Figure \ref{fig:blam} shows the 3D maps generated with BLAM and the horizontal Velodyne. For comparison we include Figure \ref{fig:blam_v}, which is generated using the localization estimate from the horizontal BLAM, but only rendering the vertical Velodyne. In the future we will color all the points using the cameras and then do colored point cloud mapping.

We can make use of the trajectory evaluation shown in Figure \ref{fig:blam_error}. We see that the error is low, but looking at Figure \ref{fig:comparison} we see that it is double the value of the good Hector maps. The pointclouds in Figure \ref{fig:blam} are good and nicely the double curtain in MARS-NoLoop. 

\begin{figure*}[tb]
		\centering
		\includegraphics[width=1.0\linewidth]{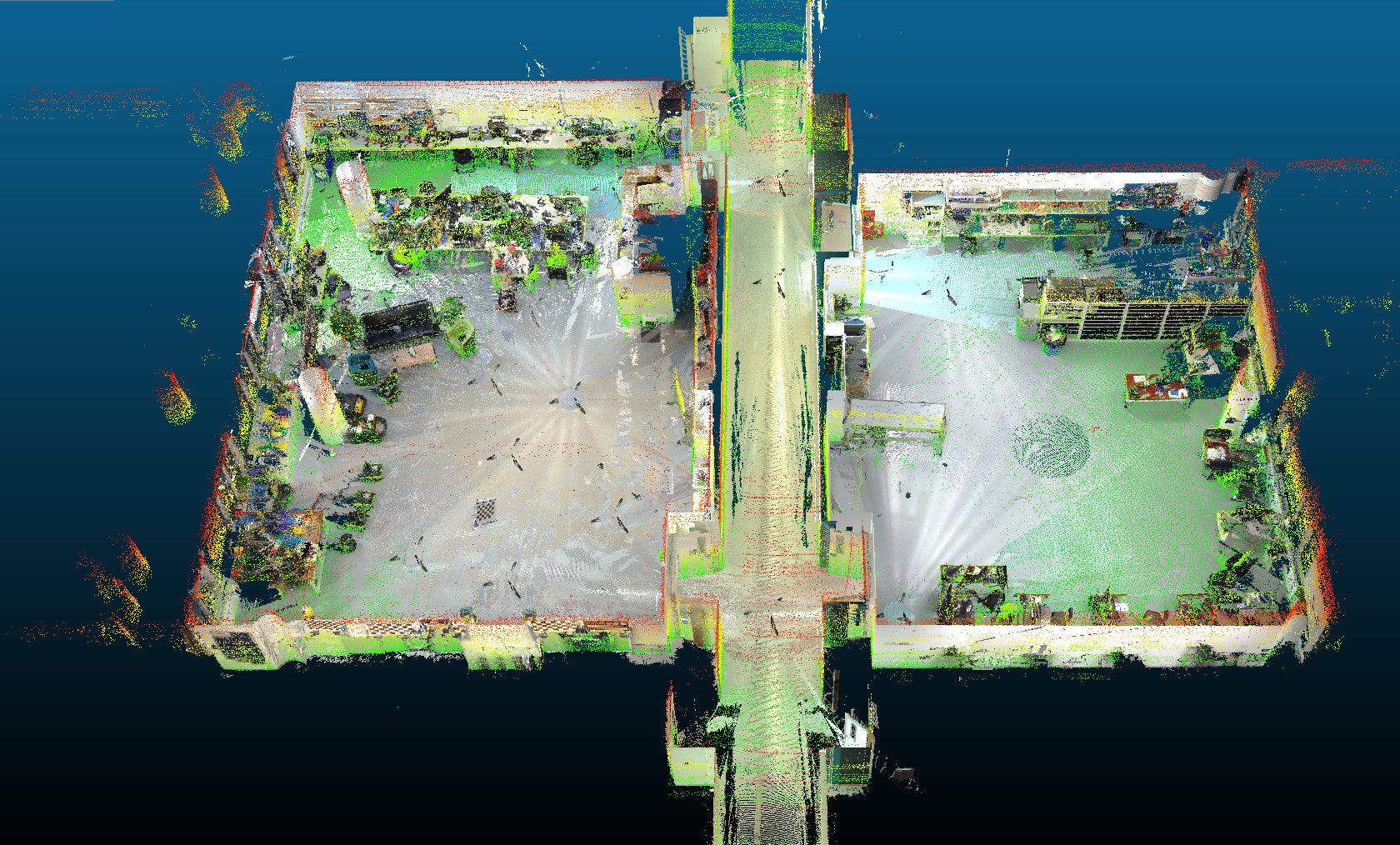}  
	\caption{Faro ground truth point cloud overlaid with BLAM MARS-Loop in cloudcompare.}
	\label{fig:cc}
\end{figure*}

We have also employed cloudcompare\footnote{\url{http://cloudcompare.org/}} for quality measurement. We register the Faro point cloud with the robot point cloud and then calculate the RMS. The result is an RMS of 0.084 with a theoretical overlap of 90\%. Figure \ref{fig:cc} shows the two point clouds overlaid.

\newcommand\cmpreSize{0.31}

\begin{figure*}[tb]
	\centering
	\includegraphics[width=\cmpreSize\linewidth]{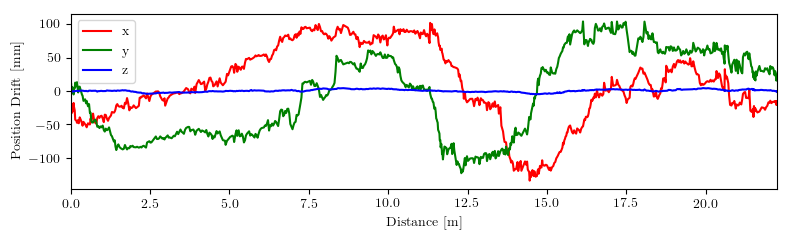}  
	\includegraphics[width=\cmpreSize\linewidth]{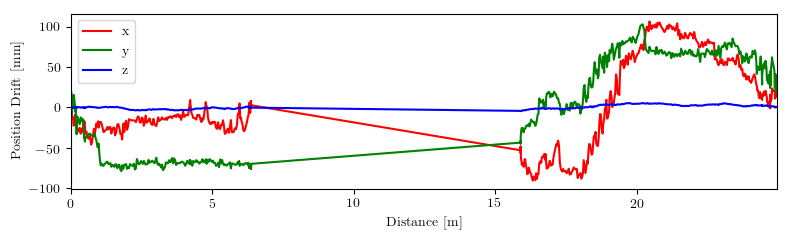}  
	\includegraphics[width=\cmpreSize\linewidth]{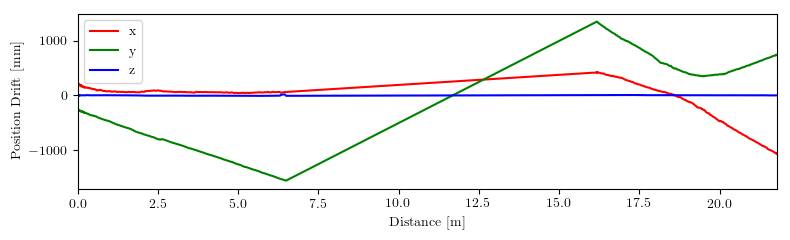}  \\ \vspace{1mm}

	\includegraphics[width=\cmpreSize\linewidth]{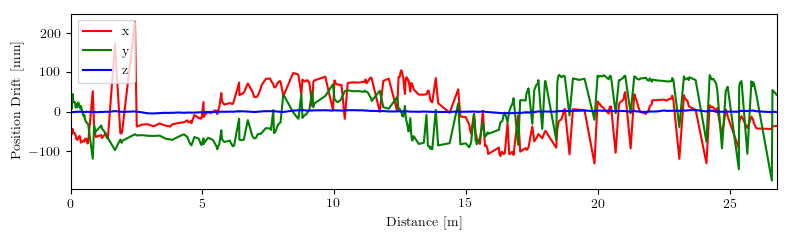}  
	\includegraphics[width=\cmpreSize\linewidth]{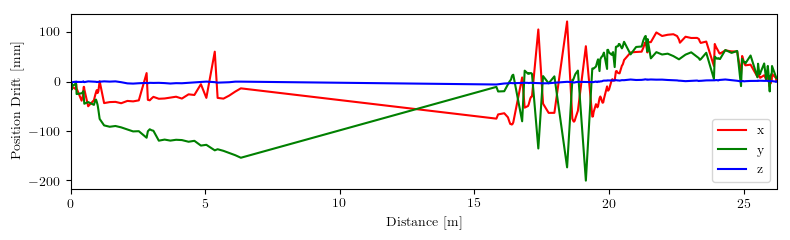}  
	\includegraphics[width=\cmpreSize\linewidth]{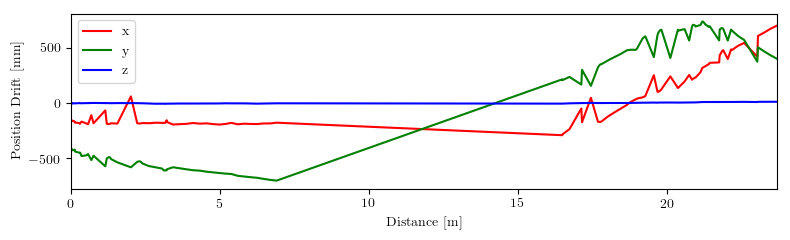}   \\ \vspace{1mm}

	\includegraphics[width=\cmpreSize\linewidth]{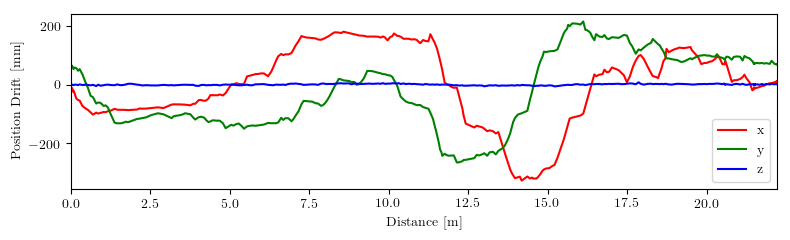}  
	\includegraphics[width=\cmpreSize\linewidth]{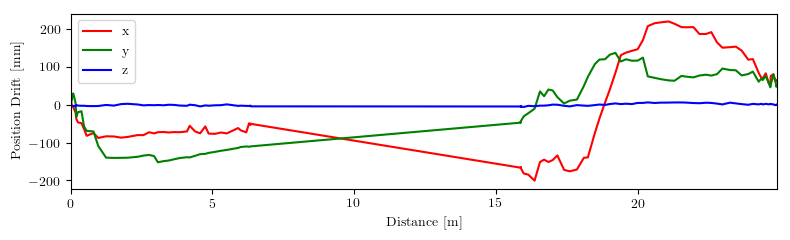}  
	\includegraphics[width=\cmpreSize\linewidth]{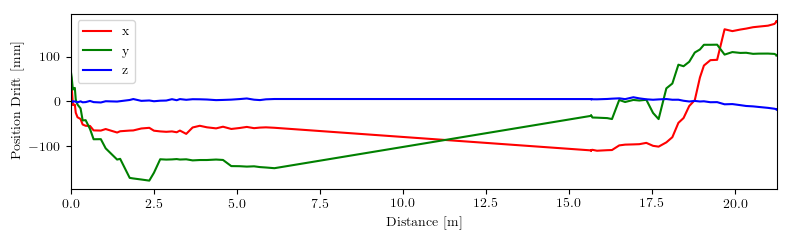}  \\ \vspace{1mm}

	\caption{The translation error in all three axes for the part of the trajectories covered by the tracking system. On the top Hector Mapping with MARS-8 (left), MARS-Loop (middle) and MARS-NoLoop (right). Below cartographer. BLAM is at the bottom.}
	\label{fig:comparison}

\end{figure*}

%

\newcommand\orgSize{0.31}

\begin{figure*}[tb]
	\centering
	\includegraphics[width=\orgSize\linewidth]{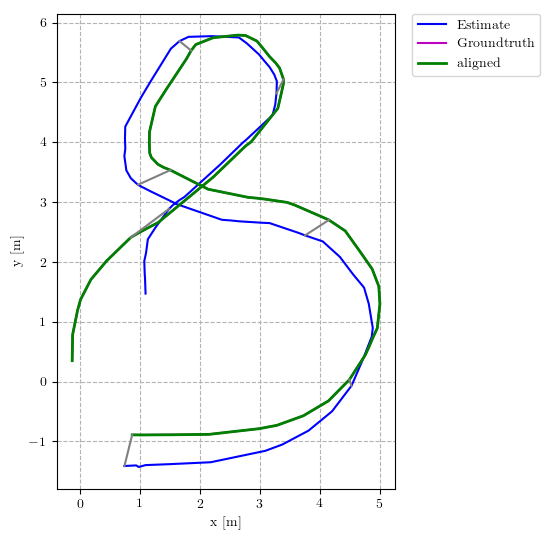}  
	\includegraphics[width=\orgSize\linewidth]{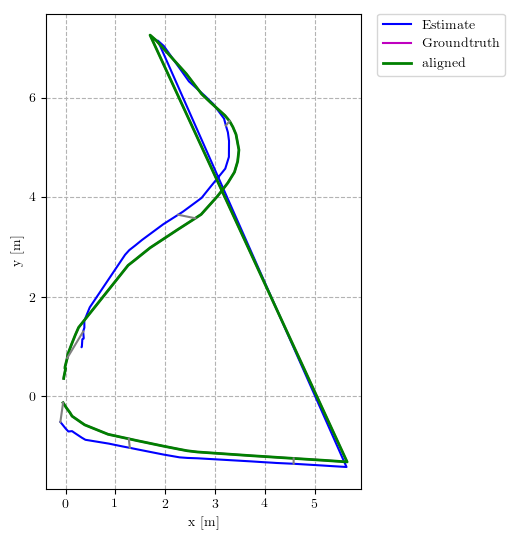}  
		\includegraphics[width=\orgSize\linewidth]{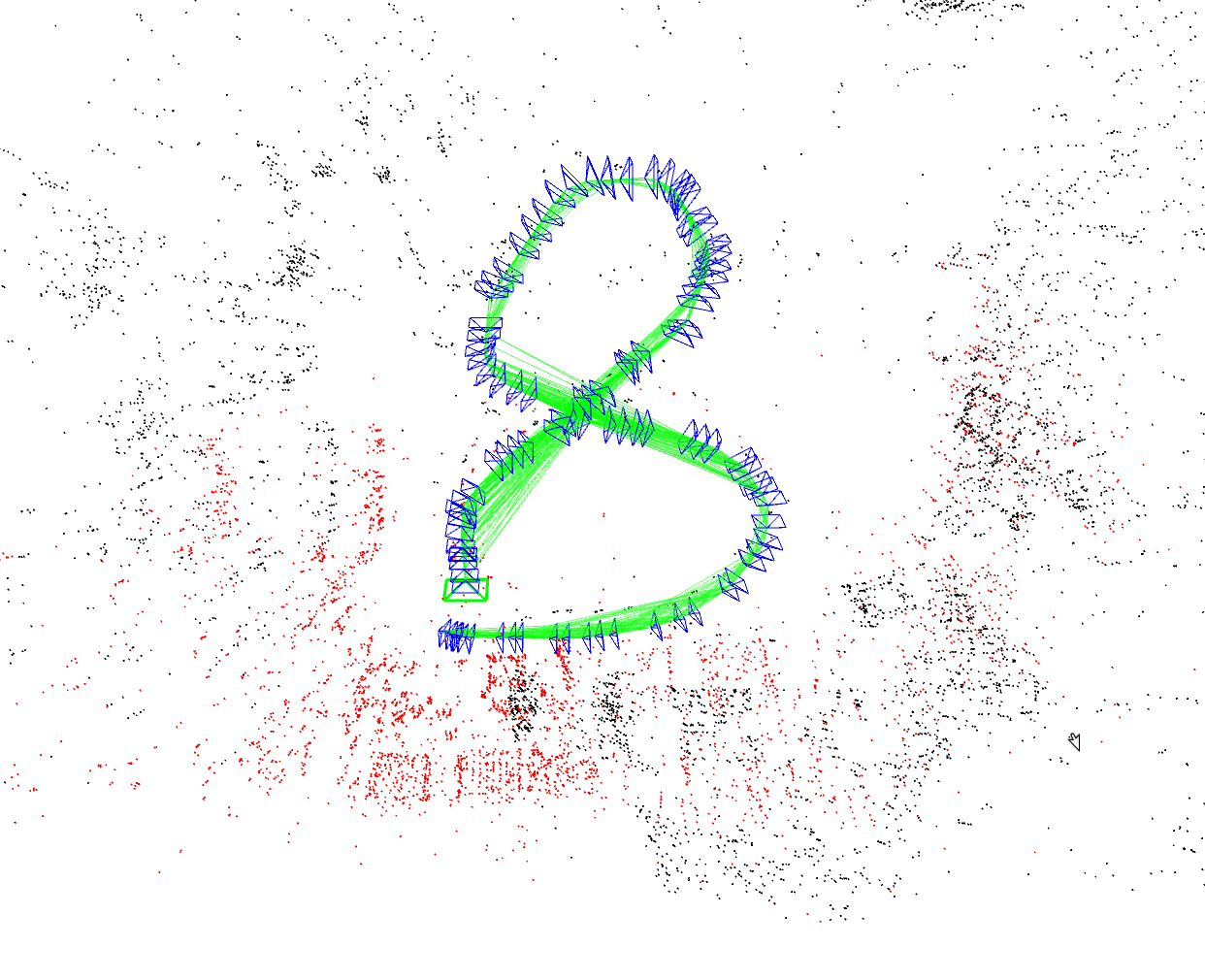}  \\ \vspace{1mm}

	\includegraphics[width=\orgSize\linewidth]{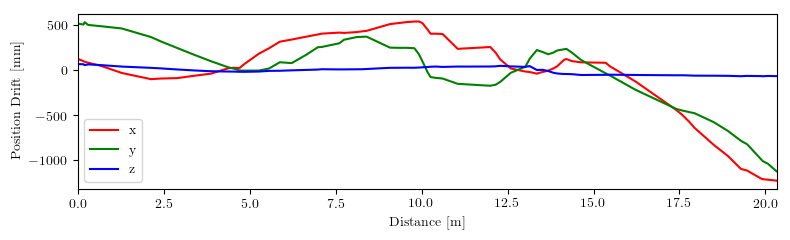}  
	\includegraphics[width=\orgSize\linewidth]{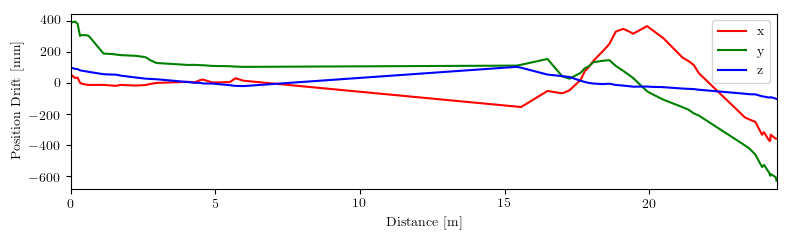} \hspace{ \orgSize\linewidth }

	\caption{Results for ORB2 visual SLAM on MARS-8 (left) and MARS-Loop (right). On the very right is the ORB2 feature cloud and trajectory for MARS-8.}
	\label{fig:orb_error}
\end{figure*}

Finally, we see the results of visual SLAM using just one camera (forward-looking on the left side) in Figure \ref{fig:orb_error}. It also shows the feature map with camera poses and pose graph. The error shown is biggest compared to the laser based SLAM algorithms. We don't show results for MARS-NoLoop, because ORB2 lost track already in the other lab. 

We will use other visual SLAM algorithms on the dataset in the future, in the hope that they will perform better. Especially when using all 9 cameras we hope to see much improved results. Another option is to explicitly make use of the four stereo camera systems.

\section{CONCLUSIONS}
\label{sec:conclusions}

In this paper we have presented our contributions in three areas: Firstly, we have described a fully hardware-synchronized advanced mapping robot for research on laser-based and visual SLAM. Secondly, we have collected three datasets for SLAM evaluation in short ranges. Thirdly, we provided repeatable evaluation procedures and compared a number of 2D, 3D and visual Simultaneous Localization and Mapping algorithms with each other, using our dataset. The results confirm the intuition, that using loop closures the error of maps can be reduced. We are also able to compare laser-based SLAM algorithms with visual SLAM algorithms and conclude, that, at least for our selection of algorithms, the laser based 3D solution outperforms visual SLAM. 

This project is still ongoing. In the final version of this paper we hope to include a few more mapping algorithms (cartographer 3D, 3D SLAM including the vertical Velodyne, other visual SLAM algorithms). We will also improve the evaluation by also employing other algorithms.
%
%


\IEEEtriggeratref{18}

\bibliographystyle{IEEEtran}
\bibliography{ref.bib,References,References_old}

\end{document}